\documentclass[10pt,letterpaper]{article}
\usepackage{acl2015,indentfirst,times,secdot}
\usepackage{xspace,empheq,fancybox,amsmath,bbm,amssymb,epsfig,subfig,syntonly,times,amsthm,graphicx} 
\usepackage{psfrag,color,bm,array}
\usepackage{url,cite,footnote,syntonly,algpseudocode}
\usepackage{verbatim,slashbox}
\usepackage[linesnumbered,ruled,vlined]{algorithm2e}

\usepackage{verbatim,multirow,diagbox,enumitem,tabu,acl2015}
\usepackage[T1]{fontenc}
\usepackage{setspace}
\usepackage[justification=justified,font={small,sf}]{caption}
\usepackage[table,xcdraw]{xcolor}
\usepackage{adjustbox}
\DeclareMathOperator*{\argmin}{arg\,min}
\usepackage[colorinlistoftodos]{todonotes}

\def\BibTeX{{\rm B\kern-.05em{\sc i\kern-.025em b}\kern-.08em
    T\kern-.1667em\lower.7ex\hbox{E}\kern-.125emX}}

\input{mysymbol.sty}

\newcommand{\hao}{\color{black}{}}
\newcommand{\KB}{\color{black}{}}

\begin{document}
	
\title{Efficient Representation for Electric Vehicle Charging Station Operations using Reinforcement Learning}
\author{\normalsize \normalfont
 Kyung-bin Kwon and Hao Zhu \\
	\normalsize  Department of Electrical and Computer Engineering \\
	\normalsize The University of Texas at Austin \\
	\normalsize  Emails: \{kwon8908kr, haozhu\}@utexas.edu}
 \maketitle

\begin{abstract}
Effectively operating an electric vehicle charging station (EVCS) is crucial for enabling the rapid transition of electrified transportation. {\KB By utilizing the flexibility of EV charging needs, the EVCS can reduce the total electricity cost for meeting the EV demand.} To solve this problem using reinforcement learning (RL), the dimension of state/action spaces unfortunately grows with the number of EVs, which becomes very large and time-varying. This dimensionality issue affects the efficiency and convergence performance of generic RL algorithms. To this end, we advocate to develop aggregation schemes for state/action according to the emergency of EV charging, or its laxity. A least-laxity first (LLF) rule is used to consider only the total charging power of the EVCS, while ensuring the feasibility of individual EV schedules. In addition, we propose an equivalent state aggregation that can guarantee to attain the same optimal policy. Using the proposed aggregation scheme, the policy gradient method is applied to find the best parameters of a linear Gaussian policy. Numerical tests have demonstrated the performance improvement of the proposed representation approaches in increasing the total reward and policy efficiency over existing approximation-based method. 
\end{abstract}
\vspace*{6pt}


\section{Introduction}
\label{sec:intro}

Electrified transportation is drastically reshaping worldwide urban mobility as a key technology to enable a future low-carbon energy society. The number of electric vehicles (EVs) continues to grow rapidly \cite{global2929international}, thanks to their high efficiency \cite{stevic2012energy} and low pollution emissions \cite{hu2021}. 
This has propelled the popularity of EV charging stations (EVCS) in metropolitan areas, as supported by significant investment in urban electricity infrastructure. 


Solving the problem of optimal operational strategies is crucial for maximizing the economic profit of EVCS owners while ensuring the quality-of-service for EV charging. {\KB In general, this problem aims to find the optimal policy for determining EV charging schedules to reduce the total electricity cost by utilizing the flexibility of EV charging needs \cite{ma2012optimal,xu2012scheduling,tang2014online,zhang2017optimal}.}
In addition, several papers have accounted for co-located renewable generation or energy storage \cite{yan2019optimized, chen2017autonomous,luo2018stochastic} or the coupling between EV traffic and electricity flow \cite{alizadeh2015optimal,he2013optimal,kzyo_trc18}. Nonetheless, one key challenge in formulating the EVCS problem lies in the randomness and uncertainty of EV arrivals and other inputs such as electricity market prices. It is possible to develop probabilistic models from actual data, such as the Gaussian distribution model of EV parking time and require demand in \cite{huang2018robust}, or the representation of the charging demand as a mixed Gaussian model to be estimated in \cite{luo2018stochastic}. Although these models have led to efficient stochastic programming approaches for the EVCS problem, they could be prone to potential modeling mismatches or fail to capture the problem dynamics therein. 

To tackle this challenge, this paper aims to develop a data-driven framework to solve the EVCS operation problem by leveraging reinforcement learning (RL) techniques \cite{sutton1998}. Using actual data samples, RL has shown some success in solving this problem with no need for stochastic modeling \cite{wan2019model, li2020constrained, wang2021reinforcement}.  
Nonetheless, most existing approaches use the original problem representation of individual EVs' status and charging action. This leads to very \textit{high} and \textit{time-varying} dimensionality for both the state and action spaces, significantly affecting the efficiency and convergence of policy search by generic RL algorithms. By transforming the EV status to the so-termed \textit{laxity} that measures the emergency level of its charging need, the work in \cite{wang2021reinforcement} has proposed to consider the total charging power across the EVCS as the action instead. Furthermore, a least-laxity first (LLF) rule has been advocated to recover individual EVs' actions from the aggregated one, which can maintain the feasibility of the charging solutions. The dimensionality issue of state space is solved by approximating the action-value function, or Q-function, which lacks approximation guarantees. 

To this end, our work has proposed a new state representation by aggregating the individual EV status into the number of EVs in each laxity group. We have analytically shown that this aggregation scheme is equivalent to the original one and thus can lead to the same optimal policy by an RL algorithm. 
The main contribution of the present paper is two-fold. First, we have developed a comprehensive representation for both the state and action spaces of the EVCS operations problem, with guaranteed equivalence to the original model. Second, the proposed representation enjoys fixed and low problem dimensions, developing an efficient algorithm to search for the optimal policy. 
Our numerical results have validated the performance improvement of the proposed state representation compared to the existing approach of Q-function approximation and suggested additional state aggregation by further grouping the higher-laxity EVs with minimal performance degradation.



The rest of paper is organized as follows. Section \ref{sec:modeling} formulates the EVCS operations problem as a Markov Decision Process (MDP). Section \ref{sec:twoways} develops the LLF-based action reduction and our proposed equivalent state aggregation to deal with dimensionality issues. Based on this, Section \ref{sec:RL} presents the reinforcement learning approach using policy gradient and linear Gaussian policy parameterization. Numerical tests using real-world data are studied in Section \ref{sec:simulation} to demonstrate the performance improvement of the proposed algorithm, and the paper is wrapped up in Section \ref{sec:conclusion}.


\section{System Modeling}
\label{sec:modeling}

\begin{figure}[t]
	\centering
	\includegraphics[width=\linewidth]{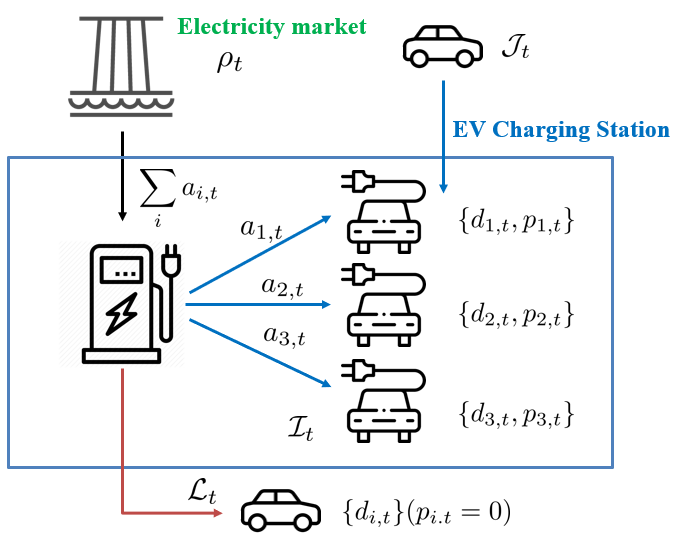}
	\caption{System Model of EV Charging Station}
	\label{fig_evcs}
\end{figure}

Consider the operations of an EV charging station (EVCS) as depicted in Fig.~\ref{fig_evcs} over the time period $\mathcal T = [0,\ldots,T]$. For each time $t\in \ccalT$, let $\mathcal{I}_t$ denote the set of parked EVs, with $\ccalJ_t$ and $\ccalL_t$ denoting the sets of arriving and departing EVs, respectively. Hence, the set of EVs is updated by $\mathcal{I}_{t+1} = (\mathcal{I}_t \cup \mathcal{J}_{t+1}) \backslash \mathcal{L}_{t+1}$, thus time-varying. Upon the arrival of EV $i\in\ccalJ_t$, its remaining demand $d_{i,t}$ and parking time $p_{i,t}$ are determined by the owner. The goal of EVCS operations is to determine the charging action $a_{i,t}$ for every parked EV $i \in \ccalI_t$, based on the real-time electricity prices $\{\rho_t\}$ received from the market operator. Each EV's status is updated according to the $\{a_{i,t}\}$ sequence, until its departure time $\tau\in\ccalT$ such that either $d_{i,\tau}=0$ or $p_{i,\tau}=0$. {\KB For simplicity, all EVs are assumed to have the same charging power, with the possibility of extension to different charging rates as analyzed in \cite{SLLF}.} 
In addition, this work assumes the charging actions will ensure each EV to be fully charged before departure; i.e., the departure time $\tau$ is the first slot with $d_{i,\tau}=0$. This assumption is reasonable because the EVCS can always increase the total charging budget to meet all EV demands. In future, we will extend it to the general case of non-fully charged EVs by introducing a penalty cost. 

This work aims to develop efficient reinforcement learning (RL) algorithm for the EVCS operation problem. To this end, we model it as a Markov Decision Process (MDP) \cite[Ch.~3]{sutton1998} denoted as a tuple $(\mathcal{S}, \mathcal{A}, \mathcal{P}, \mathcal{R}, \gamma)$, as detailed here.

\textbf{State space} $\mathcal{S}$ contains the set of feasible values for both the EV-internal and external status variables. This includes the remaining demand and parking time for each EV, as well as the electricity market price $\rho_t$. Hence, the state per time $t$ is given by $s_t = [\rho_t, \{d_{i,t}, p_{i,t}\}_{i \in \mathcal{I}_t}]$. 

\textbf{Action space} $\mathcal{A}$ includes the set of decisions that the active EVs can take. Without loss of generality (Wlog), consider a simple binary decision rule for each EV as given by $a_{i,t} \in \{ 0~\text{(do nothing)}, 1~ \text{(charge)}\}$. It can be extended to a multi-level charging rate with $|\ccalA| >2$ or a continuous charging action. For simplicity, this paper focuses on the case of binary action. 

\textbf{Transition kernel} $\mathcal{P}: \mathcal{S} \times \mathcal{A} \times \mathcal{S} \rightarrow [0,1] $ captures the system dynamics under the Markov property {\cite[Ch.~3]{sutton1998}}. In the case of stochastic electricity market prices, we assume Pr$(\rho_{t+1}|\{\rho_\tau\}_{\tau=1}^t)=$ Pr$(\rho_{t+1}|\rho_t)$. This is reasonable since the market price has short-term memory {\cite{tryggvi2012forecasting}}. A longer memory is possible too; such as the prices that follow Pr$(\rho_{t+1}|\{\rho_\tau\}_{\tau=1}^t)=$ Pr$(\rho_{t+1}|\rho_t,~\rho_{t-1})$. In this case, both $\rho_t$ and $\rho_{t-1}$ are included as the part of the state per time $t$ to satisfy the Markov transition property.

In addition, the EV status is updated according to the charging action in a deterministic fashion. For simplicity, let $d_{i,t}$ and $p_{i,t}$ denote the number of time slots for EV $i$ to attain full charging and stay parked at time $t$, respectively. This way, their transitions are given by 
\begin{align}
d_{i,t+1} = d_{i,t}-a_{i,t},~\text{and}~p_{i,t+1} = p_{i,t}-1. \label{eq:dp}
\end{align}
This update rule also holds for general action spaces if $a_{i,t}$ is not binary.

\textbf{Reward function} $\mathcal{R}: \mathcal{S} \times \mathcal{A} \rightarrow \mathbb{R}$ indicates the instantaneous reward used for defining the optimal actions. Wlog, assume all EVs have the same charging rate and thus the reward related to the total charging cost in time $t$ is given by $r_t (s_t,~a_t)= - \rho_t (\sum_{i \in \mathcal{I}_t} a_{i,t})$. 
The reward objective can also consider other economic factors such as peak demand reduction and load shaping benefits. 
 
\textbf{Discount factor} $\gamma \in (0,1]$ is a constant to accumulate the total reward along the time horizon. Smaller $\gamma$ values imply that future rewards are less important than current ones at a discounted rate \cite[Ch.~3]{sutton1998}. For this finite time-horizon problem, $\gamma = 1$ will be used for simplicity. 

For the MDP-based model, we can formulate the EVCS operation problem. The goal is to find the optimal policy $\pi$ for mapping $a_t\sim \pi(s_t)$ with $s_t$. To simplify the policy search, we are particularly interested in the set of parameterized policies given by $\pi_\mu (\cdot) = \pi (\cdot; \mu)$, which optimizes over parameter $\mu$ as given by
\begin{align}
\max_{\mu} J(\mu)= &V^{\pi}(s_0) \nonumber\\
:= & \mathbb{E}_{a_t \sim \pi_\mu (s_t), \ccalP} \left[ \sum_{t=0}^{T} \gamma^{t} r_t (s_t,~a_t) \Big | s_0 \right]  
 \label{obj_parameter}
\end{align}
where $V^{\pi}(s_0)$ denotes the value function for given initial state $s_0$. 
The formulation \eqref{obj_parameter} allows for adopting popular RL algorithms. {\KB The parameterized model and problem set-up will be discussed with more details in Section \ref{sec:RL} along with the policy gradient (PG) solution method \cite{sutton2000advances}.} Notably, the dimensions of state and action in \eqref{obj_parameter} can be very high and are time-varying, making it challenging to search for an effective policy using RL. The following section will develop efficient state/action representation for the EVCS problem. 


\section{Efficient MDP Representation}
\label{sec:twoways}

Solving the MDP problem is challenged by the state/action representations of high dimension and time-varying. As the policy maps from state to action, the number of parameters in $\mu$ would grow with both state/action dimensions. This increasing rate would significantly slow down the search for an effective policy by generic RL algorithms. 
To tackle these issues, we propose considering the action reduction using the least-laxity first (LLF) rule and proposing an equivalent state aggregation through laxity-based grouping.

\subsection{LLF-based action reduction}
\label{sec:action}

{ We can reduce the action space to $\ccalA'$ that only consists of the total charging action $a_t = \sum_{i\in\mathcal{I}_t} a_{i,t}$.} This way, the instantaneous reward becomes $r_t = - \rho_t \cdot a_t$. To recover each $a_{i,t}$ from $a_t$, we adopt the LLF rule proposed in \cite{wang2021reinforcement} to rank the priority of EVs according to the laxity, as defined by $\ell_{i,t} := p_{i,t}-d_{i,t}$. The smaller $\ell_{i,t}$ is, the fewer flexible slots EV $i$ can use to skip charging before departure, and thus the more emergent it is at time $t$ compared to other EVs. 
If $\ell_{i,t} = 0$, or $p_{i,t}=d_{i,t}$, then EV $i$ needs to be charged throughout its remaining parking time to be fully charged before departure. The LLF rule aims to increase the flexibility of EV charging by serving the least flexible ones first.

\begin{table}[t]
	\begin{tabular}{ccccccc}
		\hline
		\multicolumn{2}{c}{\textbf{Time period}} & \textbf{$t=0$} & {\color[HTML]{FE0000} \textbf{$t=1$}} & \multicolumn{1}{l}{\textbf{$t=2$}} & \textbf{$t=3$} & \textbf{$t=4$} \\ \hline
		\multicolumn{2}{c}{$a_{t}$}              & 2              & {\color[HTML]{FE0000} 1}              & 0                                  & 2              & 0              \\ \hline
		& $d_{1,t}$   & 3              & {\color[HTML]{FE0000} 2}              & 1                                  & 1              & 0              \\
		& $p_{1,t}$   & 4              & {\color[HTML]{FE0000} 3}              & 2                                  & 1              & 0              \\
		& $\ell_{1,t}$   & 1              & {\color[HTML]{FE0000} 1}              & 1                                  & 0              & 0              \\ \cline{2-7} 
		\multirow{-4}{*}{$EV_1$}   & $a_{1,t}$   & 1              & {\color[HTML]{FE0000} 1}              & 0                                  & 1              & 0              \\ \hline
		& $d_{2,t}$   & 2              & {\color[HTML]{FE0000} 1}              & 1                                  & 1              & 0              \\
		& $p_{2,t}$   & 4              & {\color[HTML]{FE0000} 3}              & 2                                  & 1              & 0              \\
		& $\ell_{2,t}$   & 2              & {\color[HTML]{FE0000} 2}              & 1                                  & 0              & 0              \\ \cline{2-7} 
		\multirow{-4}{*}{$EV_2$}   & $a_{2,t}$   & 1              & {\color[HTML]{FE0000} 0}              & 0                                  & 1              & 0              \\ \hline
	\end{tabular}
	\caption{Two-EV example by following LLF rule.}
	\label{table_laxity_exp}
\end{table}

\begin{table}[t]
	\begin{tabular}{ccccccc}
		\hline
		\multicolumn{2}{c}{\textbf{Time period}} & \textbf{$t=0$} & {\color[HTML]{FE0000} \textbf{$t=1$}} & \multicolumn{1}{l}{\textbf{$t=2$}} & \textbf{$t=3$} & \textbf{$t=4$} \\ \hline
		\multicolumn{2}{c}{$a_{t}$}              & 2              & {\color[HTML]{FE0000} 1}              & 0                                  & 2              & 0              \\ \hline
		& $d_{1,t}$   & 3              & {\color[HTML]{FE0000} 2}              & 2                                  & 2              & 1              \\
		& $p_{1,t}$   & 4              & {\color[HTML]{FE0000} 3}              & 2                                  & 1              & 0              \\
		& $\ell_{1,t}$   & 1              & {\color[HTML]{FE0000} 1}              & 0                                  & -1              & -1              \\ \cline{2-7} 
		\multirow{-4}{*}{$EV_1$}   & $a_{1,t}$   & 1              & {\color[HTML]{FE0000} 0}              & 0                                  & 1              & 0              \\ \hline
		& $d_{2,t}$   & 2              & {\color[HTML]{FE0000} 1}              & 0                                  & 0              & 0              \\
		& $p_{2,t}$   & 4              & {\color[HTML]{FE0000} 3}              & 2                                  & 1              & 0              \\
		& $\ell_{2,t}$   & 2              & {\color[HTML]{FE0000} 2}              & 0                                  & 0              & 0              \\ \cline{2-7} 
		\multirow{-4}{*}{$EV_2$}   & $a_{2,t}$   & 1              & {\color[HTML]{FE0000} 1}              & 0                                  & 0              & 0              \\ \hline
	\end{tabular}
	\caption{Two-EV example not following the LLF rule.}
	\label{table_nolaxity_exp}
\end{table}

To demonstrate the advantage of LLF-based action recovery, we use a simple example of only two EVs in the charging station as indexed by $EV_1$ and $EV_2$, respectively. A total horizon of $T=4$ is considered, and a possible initial state is given in Table~\ref{table_laxity_exp}. 
Under a given sequence of total charging actions $a_t$, Table~\ref{table_laxity_exp} lists the individual charging actions following the LLF rule, while Table~\ref{table_nolaxity_exp} shows one case of not following it.   
In Table ~\ref{table_nolaxity_exp}, $EV_2$ is charged at $t=1$ instead of $EV_1$ even though $\ell_{2,1} > \ell_{1,1}$. As a result, $EV_1$ is not fully charged at the end, while the total charging sequence $\{a_t\}$ has led to both EVs being fully charged in Table~\ref{table_laxity_exp}. This comparison points out the importance of having the LLF rule in disaggregating the total $a_t$. 

With the given total charging budget $a_t$, \textbf{Algorithm~\ref{alg:llf}} demonstrates a procedure for selecting EVs to charge at time $t$ according to the LLF rule.
\begin{algorithm}[t]
\SetAlgoLined
\caption{Least-laxity first (LLF) rule}
\label{alg:llf}
\DontPrintSemicolon
{\bf Inputs:} Total charging power $a_t$, the set of EVs in $\mathcal{I}_t$ along with their remaining demand $d_{i,t}$ and parking time $p_{i,t}$.\; 
{\bf Initialize:} the allocated charging budget $a = 0$. \;
Compute the laxity for each EV $i\in\mathcal{I}_t$ as $\ell_{i,t} := p_{i,t}-d_{i,t}$ and set $a_{i,t} = 0$ to indicate that it is not yet selected for charging.\;
\While{$a \leq a_t$}{
Search for the least-laxity EV $k = \argmin_{i: a_{i,t} = 0 } \ell_{i,t}$ from the remaining unchosen EVs by arbitrarily breaking the tie if there is any.\;
Set $a_{k,t} = 1$.\; 
$a \leftarrow a+1$
}
\end{algorithm}
The LLF based action reduction allows to recover feasible individual EV schedules, as shown in \cite{wang2021reinforcement} and restated here for completeness. 

\begin{proposition} \label{prop:action}
	If the EVCS total charging schedule $\{a_t\}_{t\in\ccalT}$ is feasible, i.e., there exist corresponding feasible charging schedules for individual EVs that ensure each EV to be fully charged before departure, then the LLF procedure in \textbf{Algorithm~\ref{alg:llf}} can produce such a feasible charging schedule for all the EVs. 
\end{proposition}

{ Instead of formally showing Proposition \ref{prop:action}, we provide some intuition behind it. For given $\{a_t\}_{t\in\ccalT}$, if there exist corresponding feasible individual EV schedules that do not follow the LLF rule, then we can transform the latter to feasible individual schedules that follow the LLF rule. Consider an arbitrary  feasible EV schedule $\{a_{i,t}\}_{i}$ for each EV $i$ that corresponds to the given $\{a_t\}_{t\in\ccalT}$, i.e., it holds that $\sum_i  a_{i,t}=a_t$ at every $t\in\ccalT$. If the former does not follow the LLF rule, then there exist two EVs, say $j$ and $k$, that violate the LLF rule at certain time $t'$. Specifically, we have $a_{j,t'} = 1$, and $a_{k,t'} = 0$ with the laxity $\ell_{j,t'} > \ell_{k,t'}$. The feasibility implies that $\ell_{j,t} \geq 0$ and $\ell_{k,t} \geq 0$, $\forall t \in \ccalT$. Hence, let us switch the charging for those two EVs at time $t'$, i.e., instead we pick EV $k$ to charge by setting $a_{j,t'} = 0$, and $a_{k,t'} = 1$. First, this switch does not change the total charging action. Second, as $\ell_{j,t'} > \ell_{k,t'}$ at time $t'$, this change still ensures feasibility or that the laxity values are always non-negative throughout the horizon $\ccalT$. Hence, this example shows that by following the LLF rule, one can always recover the feasible individual EV schedules. Detailed proof for this result can be found in \cite{wang2021reinforcement}.} 

\subsection{Equivalent state aggregation}
\label{sec:state}

In addition to action reduction, we also develop a state aggregation scheme to address the variable and high dimensionality issues of $\ccalS$. We pursue the ideal \textit{equivalent state aggregation} \cite{Givan2003} such that the new state space $\ccalS'$ can maintain the necessary information in $\ccalS$. The aggregation needs to ensure that both $\ccalS$ and $\ccalS'$ attain the same value functions $V^\pi(\cdot)$ and thus the same optimal policies for any given action in $\ccalA'$. Two conditions need to hold \cite{Givan2003}, as defined here. 

\begin{definition}
A state aggregation scheme $\ccalS \rightarrow \ccalS'$ satisfies \textbf{reward homogeneity} if for any pair of original states $\{s_t^{(i)}, s_t^{(j)}\}$ that will be aggregated into the same new state in $\ccalS'$, it holds that
\begin{align}
r_t(s^{(i)}_t, a_t) = r_t(s^{(j)}_t, a_t), \;\forall a_t \in \ccalA'
\end{align}
\end{definition}

\begin{definition}
A state aggregation scheme $\ccalS \rightarrow \ccalS'$ satisfies \textbf{dynamic homogeneity} if for any pair of original states $\{s_t^{(i)}, s_t^{(j)}\}$ that will be aggregated into the same new state in $\ccalS'$, it holds that 
\begin{align}
 \mathrm{Pr}(s_{t+1} | s_t = s^{(i)}, a_t) 
&= \mathrm{Pr}(s_{t+1} | s_t = s^{(j)}, a_t), \nonumber \\
~&\forall s_{t+1}\in\ccalS,~a_t\in\ccalA'
\end{align}
\end{definition}
To achieve these homogeneity conditions, we propose to aggregate the parked EVs at time $t$ into the number of EVs for every integer-valued laxity level in $[0,~L]$, where $L := \max_{i,t} \ell_{i,t}$ denotes the maximally possible laxity level at the EVCS. Note that as all EVs are assumed to be fully charged before departure, the laxity is always non-negative with the minimum equal to zero. Upon determining each EV's laxity as in Section~\ref{sec:action}, we define the aggregated state 
\begin{align}
	s'_t = [\rho_t, n^{(0)}_t, n^{(1)}_t, \cdots, n^{(L)}_t] \in \ccalS' \label{eq:agg}
\end{align}  
with $n^{(\ell)}_t$ denoting the number of EVs with laxity equal to $\ell$.  In order to show the new MDP is equivalent to the original one, let us consider the two homogeneity conditions. First, the reward homogeneity is easily satisfied as $r_t = -\rho_t  a_t$ is not affected by the aggregation. Second, dynamic homogeneity also holds due to the LLF rule for action reduction. 
Upon recovering the individual EV actions $\{a_{i,t}\}_i$ from $a_t$, the original MDP transition in \eqref{eq:dp} states that $(d_{i,t+1}, p_{i,t+1})= (d_{i,t}-a_{i,t}, p_{i,t}-1)$ for each $i \in \ccalI_t$. For the new MDP through aggregation, the state transition instead depends on the allocation of $a_t$ to each subset of EVs of the same laxity. Specifically, if $a_{i,t} = 1$ or EV $i$ is charged at time $t$, its laxity stays unchanged as $\ell_{i,t+1} = \ell_{i,t}$. Otherwise, its laxity is reduced by one as $\ell_{i,t+1} = \ell_{i,t}-1$. We can update the subset of EVs with laxity $\ell$ for time $t+1$ based on those of laxity $\ell$ at time $t$ that are charged, those of laxity $(\ell+1)$ that are not charged, along with the new arrival or departure at time $(t+1)$, as given by 
\begin{align}
n^{(\ell)}_{t+1} = a^{(\ell)}_t + [n^{(\ell+1)}_t - a^{(\ell+1)}_t] + x^{(\ell)}_{t+1} - y^{(\ell)}_{t+1},~\forall l \label{agg_trans}
\end{align}
where $a^{(\ell)}_t$ denotes the number of EVs of laxity $\ell$ that are charged in time $t$, while $x^{(\ell)}_{t+1}$ and $y^{(\ell)}_{t+1}$ representing the number of EVs of laxity $\ell$ that arrive/depart at time $(t+1)$, respectively. Similar to the LLF-based action recovery in Section~\ref{sec:action}, we allocate the total charging budget $a_t$ into each $a^{(\ell)}_t$ in an ordered fashion, as given by
\begin{align}
a^{(\ell)}_t = \min\left\{n^{(\ell)}_t, \min\left\{ a_t-
\sum_{\ell=0}^{\ell-1}a^{(\ell)}_t, 0\right\} \right\},~\forall \ell. \label{agg_a}
\end{align}
Basically, starting from the smallest laxity level $\ell=0$, we set $a^{(\ell)}_{t}=n^{(\ell)}_{t}$ until the total charging budget is met. 
Based on the two homogeneity conditions, we can formally establish the following proposition using the result from \cite{Givan2003}. 

\begin{proposition}\label{prop:state}
	Consider the original MDP $(\ccalS,\ccalA',\ccalP, \ccalR, \gamma)$ and the new MDP $(\ccalS',\ccalA',\ccalP, \ccalR, \gamma)$. If $\ccalS'$ is aggregated through $s'_t = [\rho_t, n^{(0)}_t, n^{(1)}_t, \cdots, n^{(L)}_t]$ with the transition following \eqref{agg_trans} and \eqref{agg_a}, then it satisfies both reward homogeneity and dynamic homogeneity and thus the two MDPs are equivalent. 
	As a result, the new MDP through aggregation can be used to obtain the optimal policies (determine the optimal actions) that are equivalent to the original ones. 
\end{proposition}

By guaranteeing the equivalence of the two MDPs, the aggregation maintains the same value function for any initial state as mentioned earlier. Hence,
the optimal policy obtained by an RL algorithm for the new MDP would be the same as the original one. This state aggregation scheme can efficiently search for the best $\pi(\cdot)$, at no sacrifice of optimality.
Note that the state aggregation can be further simplified in practice by merging the higher-laxity groups. If the maximum laxity $L$ is very large, the equivalent aggregation can still be of quite large dimension. Our numerical experiences suggest that the groups of higher laxity values play similar role in determining the optimal action, as the LLF rule implies that the recovered action (or the transition) would mostly depend on the groups of smaller laxity values. Hence, we can cap the number of laxity groups at a value $L_{\max} < L$ such that $n^{(L_{\max})} =\sum_{\ell \geq L_{\max} } n^{(\ell)}$. Although this further simplification may not be equivalent, it can be effective in addressing the immense value of laxity in practice.

\section{Learning the Optimal Policy}
\label{sec:RL}

{\KB The proposed efficient MDP representation has successfully handled the dimensionality issue for state/action, and will be leveraged to efficiently solve for the optimal policy $\pi$ in \eqref{obj_parameter} using general RL algorithms. Recall that the unknown policy $\pi (\cdot)$ is assumed to follow certain parameterized model, and thus the problem is to find the optimal parameter $\mu$ for the mapping $a \sim \pi_\mu(s')$. The choice of parameterized model can affect the performance of RL algorithms. Without loss of generalizability,} we consider a simple model of $\pi_\mu$ and adopt the policy gradient (PG) method \cite{sutton2000advances} to search for the best $\mu$. We use the linear Gaussian policy \cite{doya2000reinforcement}, which is popular for continuous spaces, as defined by the conditional distribution
\begin{align}
	a \sim \pi_\mu(s') = \pi_\mu(a|s') = \ccalN(\mu_s^\top s' + \overline{\mu}, \sigma^2) \label{eq:pi_mu}
\end{align}
with parameter $\mu = [\mu_s;~\overline{\mu}]$ relating $s'$ to the mean for the Gaussian distributed action $a$. The variance $\sigma^2$ can be either part of the parameter or pre-determined as exploration noise. Equivalently, the random action in \eqref{eq:pi_mu} can be simply generated by the following \textit{linear policy}
\begin{align}
	a = \mu_s^\top s' + \overline{\mu} + e \label{eq:pi_mu_lin}
\end{align}
where the additive noise $e\sim \ccalN(0,\sigma^2)$.  Using \eqref{eq:pi_mu}, the
total reward function in \eqref{obj_parameter} now becomes 
\begin{align}
J(\mu) = \int_{a\in \ccalA'} \pi_{\mu}(a | s') Q_{\mu}(s',a) \text{d}a,
\end{align}
with the Q-function, or action-value function, given by 
\begin{align}
&Q^{\pi}(s',a) \nonumber \\ 
&:= \mathbb{E}_{a_t \sim \pi_\mu (s_t), \ccalP} \left(\sum_{t=0}^T \gamma^t r_t |s_0 = s',a_0 =a\right). \label{q_function}
\end{align}

{\KB Before discussing the PG method, it is worth mentioning that other choices of $\pi_\mu$ can be readily applied as well. For example, one can use a nonlinear neural network to parameterize the Q-function, known as the Deep Q-Network (DQN) approach {\hao \cite[Ch.~20]{sutton1998}}. The proposed state/action aggregation would be powerful for accelerating these nonlinear policy based RL methods too, which can be greatly affected by the dimensionality issue.}

To maximize $J(\mu)$, we are interested to find its gradient over $\mu$ following from the \textit{log-derivative trick}, as
\begin{align}
\nabla_{\mu} J(\mu) 
&= \mathbb{E}_{a \sim \pi_\mu (s)} \left[Q^{\pi}(s',a) \nabla_{\mu} \ln \pi_{\mu}(a|s') \right].
\label{J_gradient}
\end{align}
Interestingly, this gradient computation boils down to that of the logarithmic term only, which can be easily obtained for Gaussian distribution as 
\begin{align}
	\nabla_{\mu_s} \ln \pi_{\mu}(a|s') &= \frac{a'-(\mu_s^\top s' + \overline{\mu})}{\sigma^2} s', \\
	\nabla_{\overline{\mu}} \ln \pi_{\mu}(a|s') &= \frac{a'-(\mu_s^\top s' + \overline{\mu})}{\sigma^2}.
\end{align}
To estimate this gradient, one can replace the expectation in \eqref{J_gradient} by the sample mean obtained from the trajectory $\{s'_0, a_0, s'_1, a_1, \cdots, s'_T, a_T\}$:
\begin{align}
\hat{\nabla}_{\mu} J(\mu) \propto \sum_{t=1}^T \hat{Q}_{\mu}(s'_t, a_t) \nabla_{\mu} \ln \pi_{\mu}(a_t|s_t). \label{est_gradient}
\end{align}
{\KB with the samples $\hat{Q}_{\mu}(s'_t, a_t) = \sum_{\tau = t}^T \gamma^{\tau-t} r_{\tau}(s'_\tau,a_\tau)$ estimated from the trajectory. }
Note that the time window for approximating $\hat{Q}_{\mu}(s'_t, a_t)$ decreases as $t$ increases under the finite time-horizon setting of $\ccalT$. For larger $t$ values, fewer samples are used and the scale of Q-value is expected to decrease. To cope with this issue, one can normalize the approximated Q-function by subtracting the mean and dividing it with the standard deviation of all episode rewards \cite{schulman2018highdimensional}. This can generally improve the training stability under the high variance of the policy gradient estimator.

With a given learning rate (step-size) $\alpha$, the policy gradient method uses the estimated gradient in \eqref{est_gradient} and implements the iterative gradient ascent updates of $\mu$. Per iteration $n$, the update becomes 
\begin{align}
\mu^{n+1} = \mu^n + \alpha \hat{\nabla}_{\mu} J(\mu^n),
\label{eq:PG}
\end{align}
until the parameters converge. {To improve the gradient update, we can incorporate multiple training samples, each of which will produce a gradient estimate. Accordingly, the sum (or average) of the gradients estimated from each training sample will be used for the update in \eqref{eq:PG}.}
%
\textbf{Algorithm \ref{alg:pgm}} has detailed steps for solving the proposed MDP representation under LLF-based action reduction and the equivalent state aggregation. 

\begin{algorithm}[t]
\SetAlgoLined
\caption{Optimal EVCS policy search}
\label{alg:pgm}
\DontPrintSemicolon
{\bf Hyperparameters:} discount factor $\gamma$, step-size $\alpha$, and exploration time period $T$. \;
{\bf Inputs:} the price sequence $\{\rho_t\}_{t=0}^T$, and the EV arrivals in $\{\ccalJ_t\}_{t=0}^T$ along with the initial states of EVs\; 
{\bf Initialize:} $\mu^0$ at iteration $n=0$. \;
\While{$\mu^n$ \textit{not converged}}{
Initialize $t=0$ with the original state $s_0$.\;
\For{$t=0, \cdots ,T-1$}{
		Find the aggregated state $s'_t$ using \eqref{eq:agg};\;
		Sample $a_t \sim \pi_{\mu_n} (s_t')$ using \eqref{eq:pi_mu};\;
		Use the LLF rule in \textbf{Algorithm~\ref{alg:llf}} to recover the individual EV charging actions $\{a_{i,t}\}_{i\in\ccalI_t}$; \;
		Compute the instantaneous reward $r_t$;\;
		Update the new state $s_{t+1}$ using \eqref{eq:dp}.\;
}
Use the sample trajectory to estimate gradient $\hat{\nabla}_{\mu} J(\mu^n)$ and perform the update in \eqref{eq:PG}; \;
Update iteration $n \leftarrow n+1$.
}
\end{algorithm}

\section{Numerical Tests}
\label{sec:simulation}

\begin{figure}[t]
	\centering
	\includegraphics[width=\linewidth]{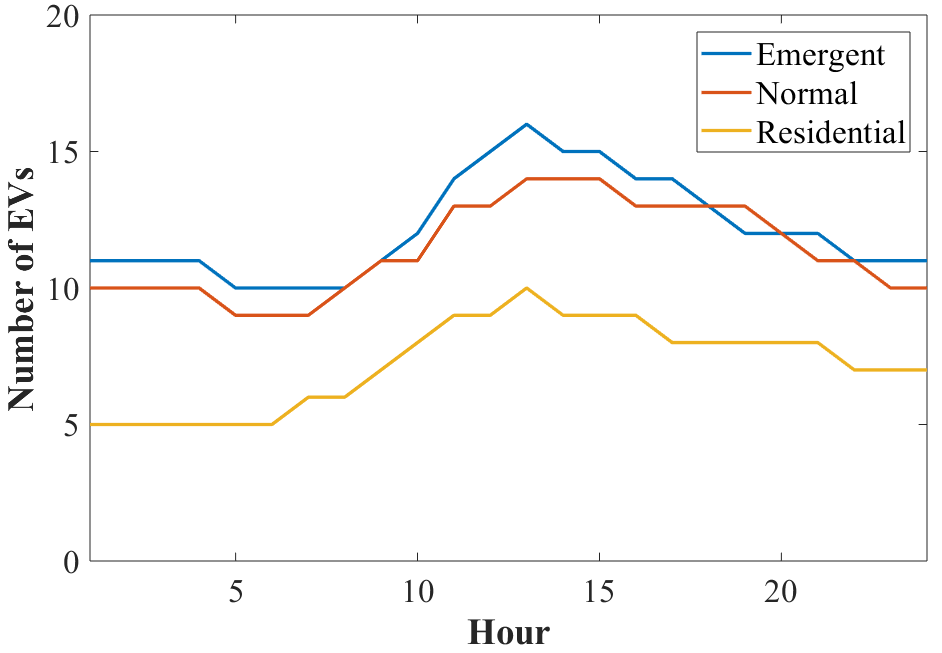}
	\caption{Hourly arrivals for the three categories of EVs during one day. }
	\label{fig_ev_arrival}
\end{figure}

We have tested the proposed \textbf{Algorithm \ref{alg:pgm}} to demonstrate the effectiveness of our new MDP representation. To set up the EVCS operation problem, we have used the hourly data of electricity market prices from the ERCOT market portal \cite{ERCOT} and the vehicle arrival data collected at the Richards Ave Station near downtown Davis, CA \cite{EV_DATA}. Three categories of EVs are considered: emergent, normal and residential uses, each having different initial demand and parking time distribution. Fig.~\ref{fig_ev_arrival} shows an example of the number of EVs in each category for a typical workday. Accordingly, the RL exploration time is the full-day period at 15-minute intervals, leading to a total horizon of $T=96$. The EV data show the maximum laxity $L=12$, and thus there are a total of 14 variables in $s'$.

We have compared \textbf{Algorithm \ref{alg:pgm}} to the existing approach by estimating an approximate Q-function in \cite{wang2021reinforcement}, denoted by \textbf{Algorithm \textit{QE}}. In \cite{wang2021reinforcement}, the same LLF-based action reduction is used while four binary feature functions approximate the Q-function to deal with the state dimensionality issue. These feature functions correspond to the charging cost or constraints on EV charging for the EVCS problem, while the total Q-function is assumed to be a linear combination of them. Hence, the RL problem becomes to estimate the best linear coefficients as the parameter based on the Bellman optimality condition for Q-function. 
Although this approach can deal with time-varying states, the approximation therein is heuristic and could be inaccurate. 

\begin{figure}[t]
	\centering
	\includegraphics[width=\linewidth]{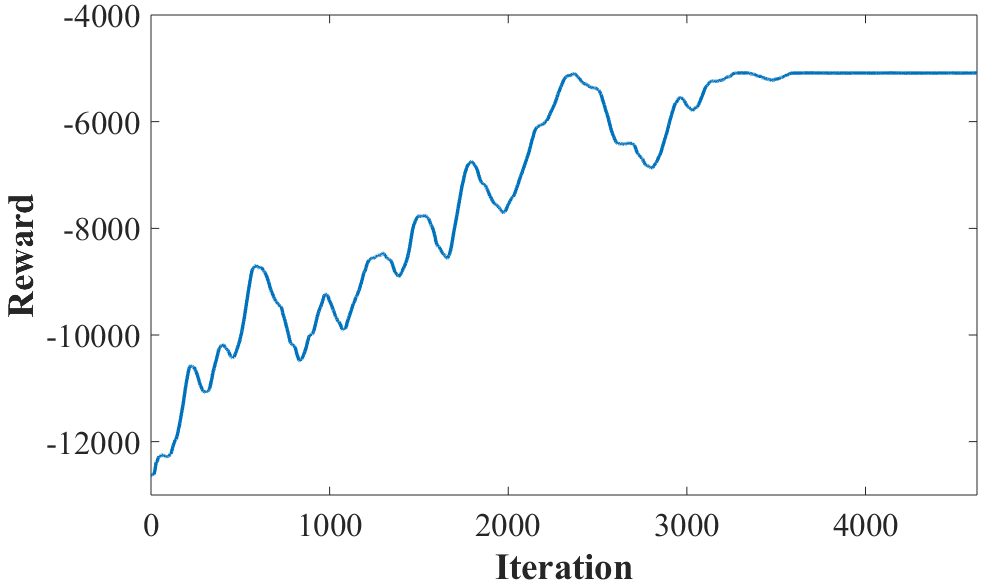} 
	\includegraphics[width=.97\linewidth]{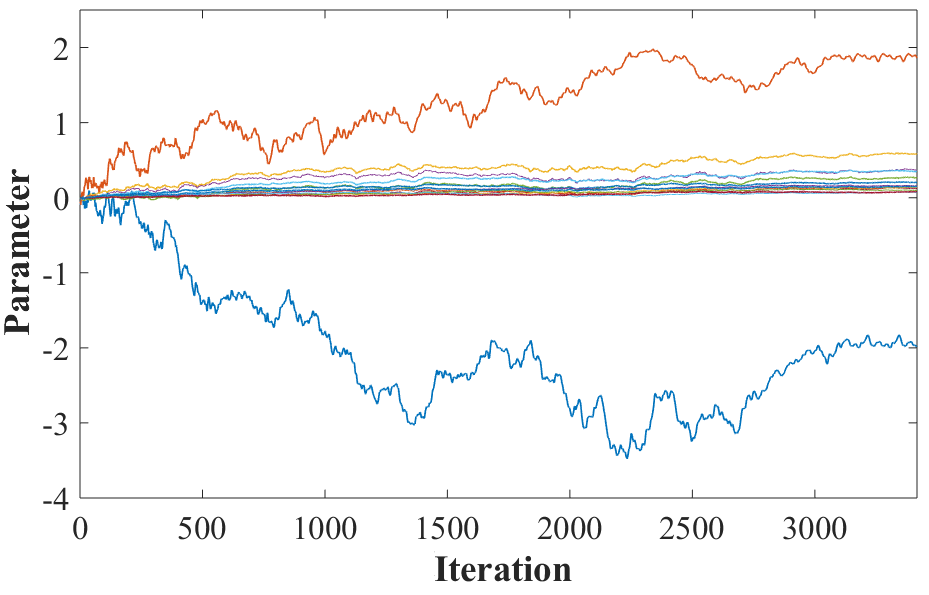} 
	\caption{(Top) The episode reward and (bottom) episode parameter values for \textbf{Algorithm \ref{alg:pgm}}.}
	\label{fig_as_converge}
\end{figure}

\begin{figure}[t]
	\centering
	\includegraphics[width=\linewidth]{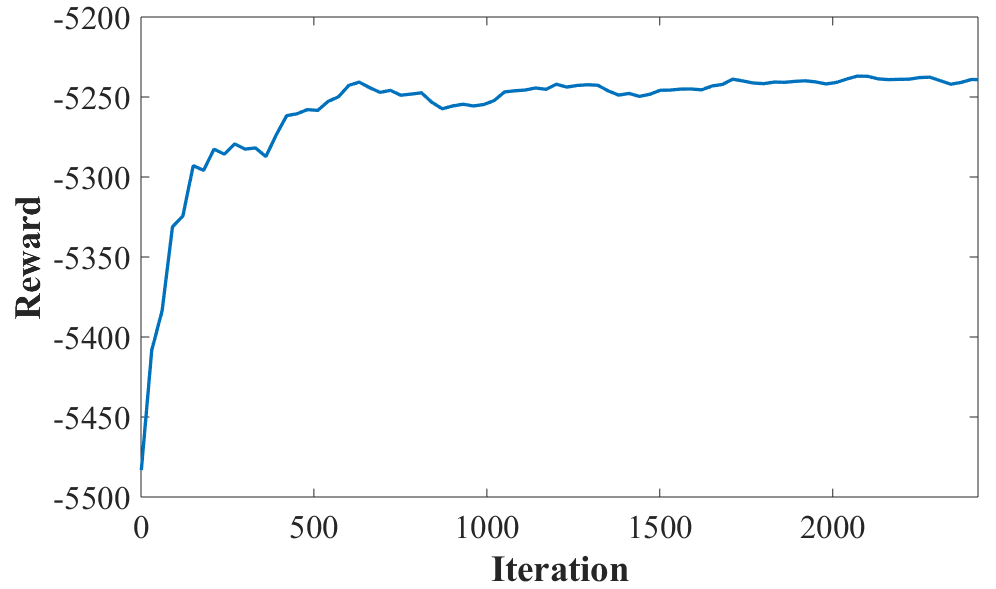} 
	\includegraphics[width=\linewidth]{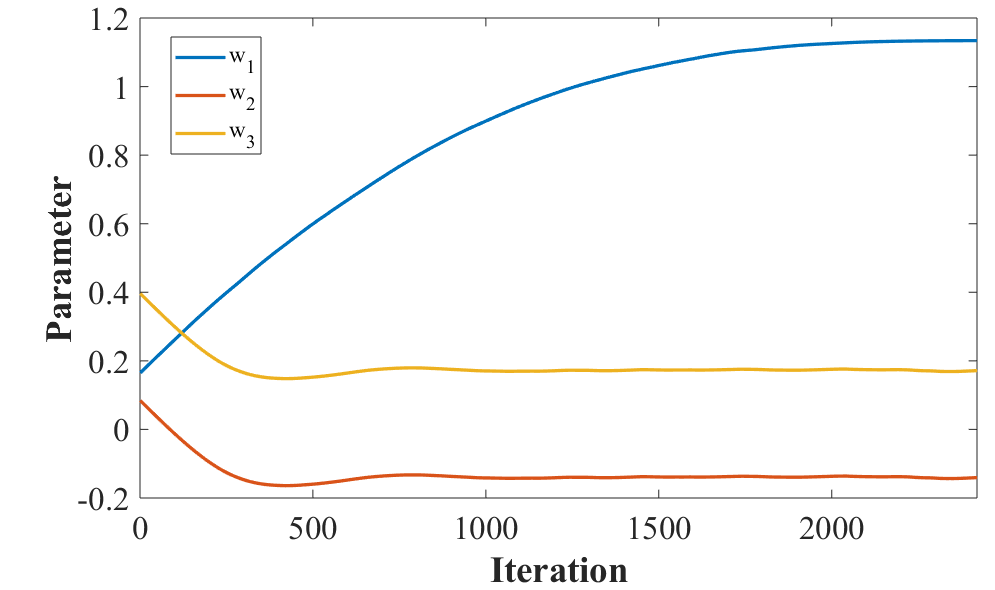} 
	\caption{(Top) The episode reward and (bottom) episode parameter values for Algorithm QE. }
	\label{fig_qe_converge}
\end{figure}

\subsection{Training results}
We have used 20 different daily profiles to train the RL algorithms. Fig.~\ref{fig_as_converge} and Fig.~\ref{fig_qe_converge} plot the episode rewards and parameter values for the proposed \textbf{Algorithm \ref{alg:pgm}} and \textbf{Algorithm \textit{QE}}, respectively. Clearly, both RL algorithms are shown to converge as rewards gradually increasing and parameter values stabilizing.  


One important observation from the episode parameter values in {Fig.~\ref{fig_as_converge}} is that they are almost zero for most states, except for state $\rho_t$ and $n^{(0)}_t$. Specifically, the negative most parameter is for $\rho_t$ as the total charging budget $a_t$ should decrease when the price is high. In addition, the positive most parameter is for $n^{(0)}_t$ as $a_t$ should increase when there are many EVs with emergent charging needs. Compared to these two parameters, the states for other laxity groups have minimal parameter values, with the parameter value decreasing at larger laxity $\ell$, as listed in Table~\ref{table_parameter}. This learning result is very reasonable as this problem depends mainly on the EVs approaching their department deadlines. As mentioned in Section \ref{sec:state}, it is possible to further reduce the number of states by merging the high-laxity EVs (larger than a threshold $L_{\max}$) into one single group. This simplification may violate the dynamic homogeneity condition, but it may not affect much the optimality of the resultant RL solution for practical systems based on this observation on minimal parameter values for high-laxity group states. 

\begin{table}[t]
	\centering
	\begin{tabular}{lclc}
		\hline
		\textbf{State} & \multicolumn{1}{c}{{Parameter}} & \textbf{State} & \multicolumn{1}{c}{{Parameter}} \\ \hline
		$\rho_t$ & -1.9735 & $n^{(6)}_{t}$  & 0.2021 \\
		$n^{(0)}_{t}$ & 1.8628 & $n^{(7)}_{t}$  & 0.1404 \\
		$n^{(1)}_{t}$  & 0.5772 & $n^{(8)}_{t}$  & 0.1386 \\
		$n^{(2)}_{t}$  & 0.3674 & $n^{(9)}_{t}$  & 0.1592 \\
		$n^{(3)}_{t}$  & 0.2651 &$n^{(10)}_{t}$  & 0.0975 \\
		$n^{(4)}_{t}$  & 0.3485 & $n^{(11)}_{t}$ & 0.0693 \\
		$n^{(5)}_{t}$  & 0.1191 & $n^{(12)}_{t}$  & 0.0797 \\ \hline
	\end{tabular}
	\caption{Parameter values obtained by \textbf{Algorithm \ref{alg:pgm}}.}
	\label{table_parameter}
\end{table}

\subsection{Testing results}

Using the two policies obtained by the RL training, we have compared their testing performances using five additional daily profiles. Table~\ref{table_comparison} lists the total reward values attained by each of the two policies for each test trajectory. Clearly, the solution by \textbf{Algorithm \ref{alg:pgm}} achieves higher total reward values, increasing those acquired by \textbf{Algorithm \textit{QE}} by around 4.15\% to 4.32\%. Thanks to the equivalent state aggregation, \textbf{Algorithm \ref{alg:pgm}} can effectively reduce the total charging cost for the EVCS. It enjoys high modeling accuracy as compared to the Q-function approximation in \cite{wang2021reinforcement}.

\begin{table}[t]
\centering
\begin{adjustbox}{width=\linewidth}
\renewcommand{\arraystretch}{2}
\begin{tabular}{lllllll}
\hline
 & \textbf{Test 1} & \textbf{Test 2} & \textbf{Test 3} & \textbf{Test 4} & \textbf{Test 5} & \textbf{Average} \\ \hline
\textbf{Alg. 2} & -5016.2 & -5022.6 & -5009.5 & -5012.8 & -5007.8 & -5013.8 \\
\textbf{Alg. QE} & -5240.1 & -5240.3 & -5234.2 & -5239.3 & -5230.6 & -5236.9 \\ \hline
\textbf{Increase (\%)} & \multicolumn{1}{c}{4.27} & \multicolumn{1}{c}{4.15} & \multicolumn{1}{c}{4.29} & \multicolumn{1}{c}{4.32} & \multicolumn{1}{c}{4.26} & \multicolumn{1}{c}{4.26} \\ \hline
\end{tabular}
\renewcommand{\arraystretch}{1}
\end{adjustbox}
\caption{Testing reward values and percentage reward increases of the solution obtained by \textbf{Algorithm \ref{alg:pgm}}, as compared to \textbf{Algorithm \textit{QE}}.}
\label{table_comparison}
\end{table}

\begin{figure}[t]
	\centering
	\includegraphics[width=\linewidth]{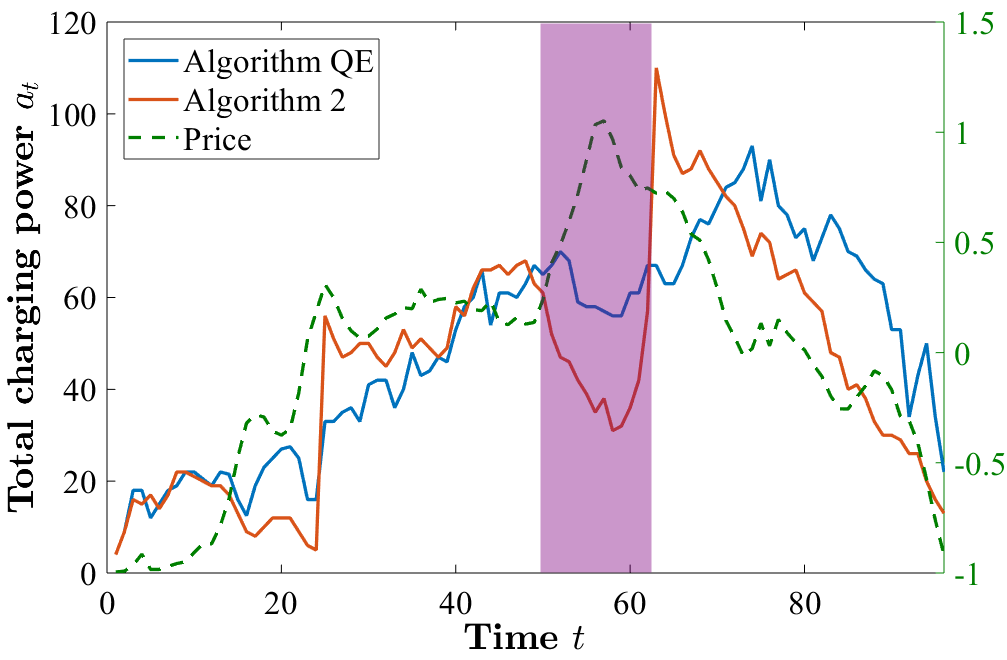}
	\caption{The daily profiles of total charging power respectively produced by Algorithms \ref{alg:pgm} and QE for one testing day as compared to the electricity market prices.} 
	\label{fig_comparison}
\end{figure}

To better illustrate the improvement of \textbf{Algorithm \ref{alg:pgm}}, Fig.~\ref{fig_comparison} plots the daily total charging action comparisons along with the electricity market price. Interestingly, \textbf{Algorithm \ref{alg:pgm}} is very sensitive to the price peaks and has chosen to dramatically reduce $a_t$. Meanwhile, \textbf{Algorithm QE} fails to reduce the charging needs over the peak-price period, {\KB as highlighted by the shaded area}. This example further verifies that our proposed EVCS operation can improve the cost performance while enjoying efficient RL solution time by considering the equivalent MDP problem.

\section{Conclusions and Future Work}
\label{sec:conclusion}

This paper has developed a practical modeling approach for the optimal EV charging station operation problem, allowing for efficient solutions using reinforcement learning (RL). To deal with the high and variable dimensions of states/actions, we propose to design efficient aggregation schemes by utilizing the EV's laxity that measures the emergency level of its charging need. First, the least-laxity first (LLF) rule has made it possible to consider only the total charging action across the EVCS, which is shown to recover feasible individual EV charging schedules if existing. 
Second, we propose aggregating the state into the number of EVs in each laxity group, which satisfies reward and dynamic homogeneities and thus leads to equivalent policy search. We have developed the policy gradient method based on the proposed MDP representation to find the optimal parameters for the linear Gaussian policy. 
Case studies based on real-world data have demonstrated the performance improvement of the proposed MDP representation over the earlier approximation-based approach for the EVCS problem. The RL parameter results imply that further state aggregation can deal with many laxity levels in practical systems at a minimal loss of optimality. 

Exciting future research directions open up regarding more general EVCS problem set-ups such as penalizing non-fully charged EVs at departure, as well as variable EV charging rate and action. The former makes it relevant to consider a constrained RL formulation that limits the number (or total demand) of unsatisfied EVs at departure or the corresponding statistical risk, {\KB following from the safe RL framework \cite{SRL}}. {\KB As for the variable charging power, it would be interesting to pursue the connection to recent work \cite{SLLF} that uses a smoothed LLF approach to deal with different charging rates.}


%
\section*{References}
\bibliography{bibliography,ecet}
\bibliographystyle{IEEEtran}
\itemsep2pt

\section*{Acknowledgments}
This work has been supported by NSF Grants 1802319, 1807097, and 1952193.

\end{document}